# Reasoning about Bayesian Network Classifiers

Hei Chan and Adnan Darwiche
Computer Science Department
University of California, Los Angeles
Los Angeles, CA 90095
{hei,darwiche}@cs.ucla.edu

## Abstract

Bayesian network classifiers are used in many fields, and one common class of classifiers are naive Bayes classifiers. In this paper, we introduce an approach for reasoning about Bayesian network classifiers in which we explicitly convert them into Ordered Decision Diagrams (ODDs), which are then used to reason about the properties of these classifiers. Specifically, we present an algorithm for converting any naive Bayes classifier into an ODD, and we show theoretically and experimentally that this algorithm can give us an ODD that is tractable in size even given an intractable number of instances. Since ODDs are tractable representations of classifiers, our algorithm allows us to efficiently test the equivalence of two naive Bayes classifiers and characterize discrepancies between them. We also show a number of additional results including a count of distinct classifiers that can be induced by changing some CPT in a naive Bayes classifier, and the range of allowable changes to a CPT which keeps the current classifier unchanged.

## 1 Introduction

A Bayesian network is a compact, graphical model of a probability distribution which assigns a probability to every event of interest [8, 6]. For example, in the medical domain, a Bayesian network can be used to compute the probability of any particular disease given the symptoms displayed by a patient.

However, when using Bayesian networks, one is often not interested in the exact probability of an event, but in whether that probability is above (or below) a certain threshold, say, .5. That is, we usually use the Bayesian network as a *classifier*, where we attempt to classify the input (e.g., patient symptoms) into a small number of, usually two, classes (e.g., whether the probability of a disease is no less than the given threshold). For example, consider the network in Figure 1, where all variables are binary. The network represents a scenario where there are three different tests for detecting pregnancy. One may use this network to classify a set of test results into whether they confirm pregnancy, depending on whether the probability of pregnancy given the results is no less than, say, .9.

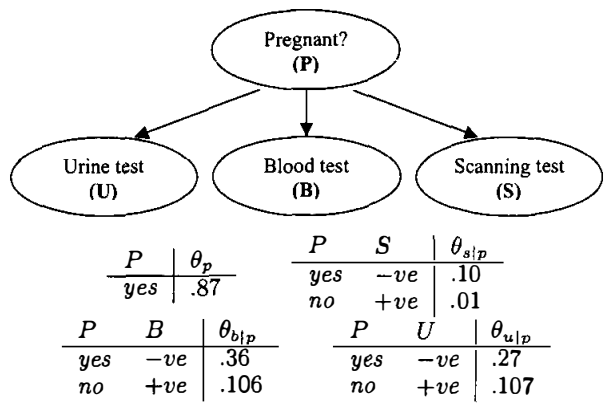

Figure 1: A Bayesian network.

The formal definition of a Bayesian network classifier is as follows. Given a Bayesian network $\mathcal{N}$, which defines the probability distribution $Pr$, we select a variable $C$, called the *class variable*, and a set of variables $\mathbf{E} = \{E_1, \ldots, E_n\}$ known as the *attributes*.[1] Each instantiation $\mathbf{e}$ of $\mathbf{E}$ is known as an *instance*. Moreover, for some probability threshold $p$, the Bayesian network can be viewed as inducing the function $F_\mathcal{N}$ which maps each instance $\mathbf{e}$ into $\{0, 1\}$ as follows: $F_\mathcal{N}(\mathbf{e}) = 1$ if $Pr(c \mid \mathbf{e}) \geq p$, and $F_\mathcal{N}(\mathbf{e}) = 0$ otherwise. The function $F_\mathcal{N}$ is called a *Bayesian network classifier* [4, 5].

---

[1]The other variables in the network are called hidden or intermediate variables. They are not mentioned and are used for modelling purposes.

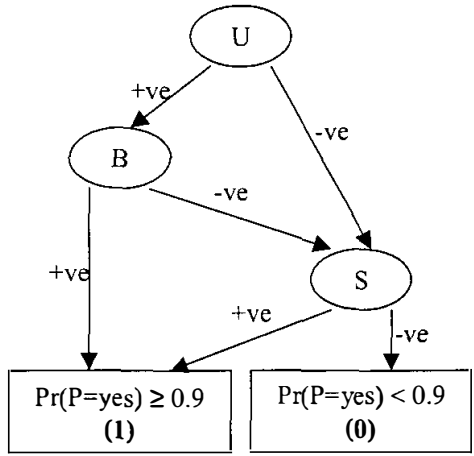

Figure 2: An ODD that represents the classifier induced by the Bayesian network in Figure 1 with probability threshold .9, with respect to variable order $(U, B, S)$.

The goal of this paper is to provide a principled approach for reasoning about Bayesian network classifiers. In particular, we are interested in answering the following type of questions:

- Given two Bayesian networks $\mathcal{N}$ and $\mathcal{N}'$, do they induce the same classifier? If not, which, and how many, instances do they disagree on?

- Given a Bayesian network $\mathcal{N}$, what are the allowable changes to some CPT in $\mathcal{N}$ which will not change the current classifier induced, $F_{\mathcal{N}}$?

These questions can be answered by enumerating all instances $\mathbf{e}$ explicitly. However, this brute–force approach is often infeasible given the exponential number of instances. Instead, we propose to build a tractable logical representation of the classifier $F_{\mathcal{N}}$, which allows us to answer the above questions in time linear in the size of the constructed representation.

The specific logical representation we propose is that of Ordered Decision Diagrams (ODDs), which are known to be tractable; see Figure 2. Although our long–term objective is to construct ODDs for general Bayesian network classifiers, we focus in this paper on the simplest, yet very common, class of naive Bayes classifiers, which are induced by naive Bayes networks.

Specifically, we start in Section 2 by defining naive Bayes classifiers, and provide the answer to the following key question: how much change can we apply to a CPT in the network without changing the current classifier induced? In Section 3, we introduce an algorithm for converting a naive Bayes classifier into an ODD, and provide an interesting asymptotic bound on its complexity. We then show in Section 4 experimental results on building ODDs for both random and real–world naive Bayes classifiers, demonstrating the scalability of our algorithm. Section 5 is dedicated to the applications of these ODDs, which are mostly enabled by the tractability of this representation. We then discuss in Section 6 our plans to extend our work beyond our proposed framework of naive Bayes classifiers. We finally close in Section 7 with some concluding remarks. Proofs of theorems are included in Appendix A.

## 2 Naive Bayes Classifiers

The simplest, yet very common, type of Bayesian network classifiers is naive Bayes classifiers [3, 7], which are induced by naive Bayes networks. A *naive Bayes network* contains the class variable $C$ as the root, with the attributes $\mathbf{E} = \{E_1, \ldots, E_n\}$ as its children. No other nodes or edges exist in the network. An example is shown in Figure 1.

To classify an instance $\mathbf{e} = \{e_1, \ldots, e_n\}$, we need to compute the conditional probability $Pr(c \mid \mathbf{e})$. However, for ease of computations, we will compute this probability in log–odds space, where its log–odds is given by $\log O(c \mid \mathbf{e}) = Pr(c \mid \mathbf{e})/(1 - Pr(c \mid \mathbf{e}))$. Given a naive Bayes network $\mathcal{N}$ where $C$ is binary,[2] if $\phi$ is an instantiation of a subset of $\mathbf{E}$, and $e_i$ is a value of an uninstantiated attribute $E_i$, we have:

$$\log O(c \mid \phi, e_i) = \log O(c \mid \phi) + \log \frac{Pr(e_i \mid c)}{Pr(e_i \mid \bar{c})}. \quad (1)$$

The *weight of evidence* $e_i$ is defined as $w_{e_i} = \log(Pr(e_i \mid c)/Pr(e_i \mid \bar{c}))$. We can now compute the value $\log O(c \mid \mathbf{e})$ using Equation 1:

$$\log O(c \mid \mathbf{e}) = \log O(c) + \sum_{i=1}^{n} w_{e_i}. \quad (2)$$

We call the value $\log O(c)$ the *prior log–odds* of $\mathcal{N}$. Therefore, a naive Bayes network is a tuple $\mathcal{N} = (C, \{E_1, \ldots, E_n\}, \log O(c), \{w_{e_i}\})$. We now formally define the *naive Bayes classifier* induced by a naive Bayes network $\mathcal{N}$ given a probability threshold.

**Definition 2.1** *Given a naive Bayesian network $\mathcal{N}$, and the threshold $\rho = \log(p/(1-p))$, where $p$ is the probability threshold, the naive Bayes classifier $F_{\mathcal{N}}^{\rho}$ is defined as follows:*

$$F_{\mathcal{N}}^{\rho}(\mathbf{e}) = \begin{cases} 1 & \text{if } \log O(c \mid \mathbf{e}) \geq \rho; \\ 0 & \text{otherwise.} \end{cases}$$

---
[2]We will make the restriction of $C$ being binary in this paper, and discuss how we will extend to the case of $C$ being non–binary in Section 6.

For example, in the naive Bayes network $\mathcal{N}$ in Figure 1, $P$ is the class variable, and $\{U, B, S\}$ are the attributes. Given the threshold $\rho = \log(.9/.1) = 2.197$, the naive Bayes classifier $F_{\mathcal{N}}^{\rho}$ determines if given an instance (a set of test results in this case), whether the probability of pregnancy is no less than .9.

We now discuss the following key question: how much change can we apply to a CPT in the network $\mathcal{N}$ without changing the current classifier induced, $F_{\mathcal{N}}^{\rho}$?

## 2.1 Changing the Prior Log–Odds

We first look at the case where we change only the CPT of the class variable $C$, and obtain a new Bayesian network $\mathcal{N}'$. This is equivalent to changing only the prior log–odds $\log O(c)$ to the new value $\log O'(c)$. Now the question is, are $F_{\mathcal{N}}^{\rho}$ and $F_{\mathcal{N}'}^{\rho}$ the same classifier? This obviously depends on the amount of change to the prior log–odds. However, the following theorem states that the amount of allowable change can be determined precisely once we know the following two values, known as margins:

- The minimum value of $\log O(c \mid \mathbf{e})$ attained by any positive instance $\mathbf{e}$:

$$\alpha = \min_{\mathbf{e}:\ F_{\mathcal{N}}^{\rho}(\mathbf{e})=1} \log O(c \mid \mathbf{e}). \quad (3)$$

- The maximum value of $\log O(c \mid \mathbf{e})$ attained by any negative instance $\mathbf{e}$:

$$\beta = \max_{\mathbf{e}:\ F_{\mathcal{N}}^{\rho}(\mathbf{e})=0} \log O(c \mid \mathbf{e}). \quad (4)$$

**Theorem 2.1** *Let $\mathcal{N}'$ be a naive Bayes network obtained from $\mathcal{N}$ by changing the CPT of the class variable $C$, such that the prior log-odds changes from $\log O(c)$ to $\log O'(c)$. The classifiers $F_{\mathcal{N}}^{\rho}$ and $F_{\mathcal{N}'}^{\rho}$ are the same iff $\log O'(c) \in [\log O(c) + \rho - \alpha, \log O(c) + \rho - \beta)$, where $\alpha$ and $\beta$ are given by Equations 3 and 4.[3] We call this interval the equivalence interval of $F_{\mathcal{N}}^{\rho}$, denoted by $I(F_{\mathcal{N}}^{\rho})$.*

Consider now the naive Bayes classifier $F_{\mathcal{N}}^{\rho}$, induced by the network in Figure 1 with threshold $\rho = 2.197$. By enumerating all instances explicitly, we find that $\alpha = 3.327$ and $\beta = .619$. Therefore, any change in the CPT of variable $P$ will keep the classifier $F_{\mathcal{N}}^{\rho}$ unchanged as long as the new prior log-odds is in the equivalence interval $I(F_{\mathcal{N}}^{\rho}) = [.772, 3.479)$. Therefore, the classifier will be unchanged as long as the new prior probability $Pr(P = yes)$ falls in $[.684, .970)$. Note that the current probability value is .87, showing that we can apply a significant change to this prior probability without changing the induced classifier. Later we will show how we can find the equivalence interval without enumerating all instances explicitly.

The maximum number of distinct naive Bayes classifiers (including the current classifier) that can be induced by changing the prior log–odds can also be counted, as given by the next theorem.

**Theorem 2.2** *The number of distinct naive Bayes classifiers (including the current classifier) that can be induced by changing prior log–odds is at most $\|\mathbf{E}\| + 1$, where $\|\mathbf{E}\|$ is the number of instances.[4]*

If all attributes are binary, this number is $2^n + 1$. For the network in Figure 1, 9 different classifiers can be induced by changing the CPT of variable $P$. Note, however, that the total number of distinct Boolean functions is $2^{2^n} = 256$ in this case.

To further illustrate Theorems 2.1 and Theorems 2.2, we will rephrase them using the mathematical notion of *equivalence class*. Given naive Bayes network $\mathcal{N} = (C, \{E_1, \ldots, E_n\}, \log O(c), \{w_{e_i}\})$, we define the set $S = \{\mathcal{N}' : \mathcal{N}' = (C, \{E_1, \ldots, E_n\}, \log O'(c), \{w_{e_i}\})\}$, i.e., $S$ contains exactly all naive Bayes networks $\mathcal{N}'$ obtained from $\mathcal{N}$ by changing only the prior log–odds (including $\mathcal{N}$). The equivalence class $[\mathcal{N}] \subset S$ is defined such that $\mathcal{N}' \in [\mathcal{N}]$ iff the classifiers $F_{\mathcal{N}}^{\rho}$ and $F_{\mathcal{N}'}^{\rho}$ are the same. Theorem 2.1 allows us to test for $\mathcal{N}' \in [\mathcal{N}]$ by verifying if $\log O'(c) \in I(F_{\mathcal{N}}^{\rho})$,[5] while Theorem 2.2 gives us a count of the number of equivalence classes that form the partition of $S$.[6]

We close this section by stressing that Theorems 2.1 and 2.2 will be crucial to our algorithm in Section 3, which converts a naive Bayes classifier into an ODD.

## 2.2 Changing all Weights of Evidence of $E_i$

We now look at the case where we change only the CPT of attribute $E_i$, and obtain the new Bayesian network $\mathcal{N}'$. This is equivalent to changing only the weight of evidence $e_i$ from $w_{e_i}$ to the new value $w'_{e_i}$ for every value $e_i$ of $E_i$. Now the question is, are $F_{\mathcal{N}}^{\rho}$ and $F_{\mathcal{N}'}^{\rho}$ the same classifier? The following theorem states this can be determined once we know the following two

---

[3] If there are no positive instances, $\alpha = \infty$, and if there are no negative instances, $\beta = -\infty$.

[4] In general, if $|X_i|$ is the cardinality of variable $X_i$, i.e., the number of possible values of $X_i$, $\|X_1, \ldots, X_k\| = \prod_{i=1}^{k} |X_i|$ is the number of instantiations of variables $X_1, \ldots, X_k$.

[5] Note that if $\mathcal{N}' \in [\mathcal{N}]$, we have $[\mathcal{N}'] = [\mathcal{N}]$ and $I(F_{\mathcal{N}'}^{\rho}) = I(F_{\mathcal{N}}^{\rho})$ by the definition of equivalence class.

[6] We note that both theorems hold not only for naive Bayes classifiers, but more generally for any Bayesian network classifier in which the attributes $\mathbf{E}$ are all descendants of the class variable $C$.

values for every $e_i$:

- The minimum value of $\log O(c \mid \mathbf{e})$ attained by any positive instance $\mathbf{e}$ such that $e_i \in \mathbf{e}$:

$$\alpha_{e_i} = \min_{\mathbf{e}:\, e_i \in \mathbf{e},\, F_\mathcal{N}^\rho(\mathbf{e})=1} \log O(c \mid \mathbf{e}). \quad (5)$$

- The maximum value of $\log O(c \mid \mathbf{e})$ attained by any negative instance $\mathbf{e}$ such that $e_i \in \mathbf{e}$:

$$\beta_{e_i} = \max_{\mathbf{e}:\, e_i \in \mathbf{e},\, F_\mathcal{N}^\rho(\mathbf{e})=0} \log O(c \mid \mathbf{e}). \quad (6)$$

**Theorem 2.3** *Let $\mathcal{N}'$ be a naive Bayes network obtained from $\mathcal{N}$ by changing the CPT of attribute $E_i$, such that the weight of evidence $e_i$ changes from $w_{e_i}$ to $w'_{e_i}$ for every value $e_i$ of $E_i$. The classifiers $F_\mathcal{N}^\rho$ and $F_{\mathcal{N}'}^\rho$ are the same iff for every $e_i$, $w'_{e_i} \in [w_{e_i} + \rho - \alpha_{e_i}, w_{e_i} + \rho - \beta_{e_i})$, where $\alpha_{e_i}$ and $\beta_{e_i}$ are given by Equations 5 and 6.*

Consider again the naive Bayes classifier $F_\mathcal{N}^\rho$, induced by the network in Figure 1 with threshold $\rho = 2.197$. If we would like to change the CPT of attribute $U$ without changing the classifier, the allowable new weights of evidence are $w'_{U=+ve} \in [.791, 3.498)$ and $w'_{U=-ve} \in [-3.294, .791)$. For example, even if we improve the reliability of the urine test by changing the probabilities $Pr(U=-ve|P=yes)$ from .27 to .1 and $Pr(U=+ve|P=no)$ from .107 to .05, the classifier will still remain unchanged.

The maximum number of distinct naive Bayes classifiers (including the current classifier) that can be induced by changing all weights of evidence of attribute $E_i$ can also be counted, as given by the next theorem.

**Theorem 2.4** *The number of distinct naive Bayes classifiers (including the current classifier) that can be induced by changing all weights of evidence of attribute $E_i$ is at most $(k+1)^b - \lfloor k/2 \rfloor^b - \lceil k/2 \rceil^b$, where $k = \|\mathbf{E} - E_i\|$ and $b = |E_i|$.*

If all attributes are binary, this number is $2^{2n-3} + 2^n + 1$. For the network in Figure 1, at most 17 different classifiers can be induced by changing the CPT of attribute $U$.

## 3 Converting a Naive Bayes Classifier into an Ordered Decision Diagram

In this section, we will introduce an algorithm that converts a naive Bayes classifier into an *Ordered Decision Diagram (ODD)*, which we will define next.

**Definition 3.1** *An Ordered Decision Diagram (ODD), with respect to variable order $(E_1, \ldots, E_n)$, is a rooted, directed, acyclic graph, with two sinks labelled with 1 and 0, called 1-SINK and 0-SINK respectively. Every node (except the sinks) in the ODD is labelled with a variable $E_i$, and for every value $e_i$ of $E_i$, there is an edge labelled with $e_i$ exiting this node. Finally, a node is labelled with $E_i$ and its child is labelled with $E_j$ only if $j > i$.*

An ODD represents a classifier $F$ with attributes $\mathbf{E} = \{E_1, \ldots, E_n\}$ as follows. Given an instance $\mathbf{e} = \{e_1, \ldots, e_n\}$, we traverse the ODD starting at the root. At a node labelled with $E_i$, we go to the child pointed by the edge labelled with $e_i$. If we reach the 1-SINK, we have $F(\mathbf{e}) = 1$, and if we reach the 0-SINK, we have $F(\mathbf{e}) = 0$. The ODD shown in Figure 2 represents the naive Bayes classifier induced by the network in Figure 1 with probability threshold .9, with respect to variable order $(U, B, S)$. If all the variables in the ODD are binary, as in this case, it is called an *Ordered Binary Decision Diagram (OBDD)* [1], a well-researched representation of boolean functions. As we will discuss in Section 5, the tractability of the ODD representation allows us to answer the questions we posed earlier in time linear in the size of the ODD.

### 3.1 Building the ODD

Suppose now that we are given a naive Bayes classifier $F_\mathcal{N}^\rho$, which is induced by the naive Bayes network $\mathcal{N} = (C, \{E_1, \ldots, E_n\}, \log O(c), \{w_{e_i}\})$ with threshold $\rho$. Our goal is to build an ODD $\mathcal{D}$ that represents $F_\mathcal{N}^\rho$, with respect to attribute order $(E_1, \ldots, E_n)$. Before we state our algorithm and its compleixty, we first explain two key observations underlying our algorithm.

First, given an instantiation $\phi = e_1, \ldots, e_k$ of the first $k$ attributes $E_1, \ldots, E_k$, we assume the node reached by the path $\phi$ from the root of ODD $\mathcal{D}$ is the root of a sub–ODD denoted by $\mathcal{D}_\phi$. A new naive Bayes network $\mathcal{N}_\phi = (C, \{E_{k+1}, \ldots, E_n\}, \log O(c \mid \phi), \{w_{e_i}\})$ can be obtained by removing attributes $E_1, \ldots, E_k$ from $\mathcal{N}$, and updating the prior log-odds to $\log O(c \mid \phi)$. Note that the output of the naive Bayes classifier $F_\mathcal{N}^\rho$ given instance $\mathbf{e} = \phi, e_{k+1}, \ldots, e_n$ can now be obtained from the new naive Bayes classifier $F_{\mathcal{N}_\phi}^\rho$, since $F_\mathcal{N}^\rho(\mathbf{e}) = F_{\mathcal{N}_\phi}^\rho(e_{k+1}, \ldots, e_n)$. Therefore, the sub–ODD $\mathcal{D}_\phi$ represents $F_{\mathcal{N}_\phi}^\rho$.

The second key observation is based on Theorem 2.1. If $\psi$ is another instantiation of attributes $E_1, \ldots, E_k$, the path $\psi$ reaches the root of the sub–ODD $\mathcal{D}_\psi$, which represents the naive Bayes classifier $F_{\mathcal{N}_\psi}^\rho$, where $\mathcal{N}_\psi = (C, \{E_{k+1}, \ldots, E_n\}, \log O(c \mid \psi), \{w_{e_i}\})$. Because $\mathcal{N}_\phi$ and $\mathcal{N}_\psi$ differ by only their prior log–odds, from Theorem 2.1, the classifiers $F_{\mathcal{N}_\phi}^\rho$ and $F_{\mathcal{N}_\psi}^\rho$ are the same iff $\log O(c \mid \psi) \in I(F_{\mathcal{N}_\phi}^\rho)$. If this is true, the two

sub–ODDs $\mathcal{D}_\phi$ and $\mathcal{D}_\psi$ are isomorphic, and we can build the ODD $\mathcal{D}$ such that the paths $\phi$ and $\psi$ reach the same node. This allows us to save space and time when building the ODD $\mathcal{D}$. The next theorem shows how we can compute the equivalence interval $I(F^\rho_{\mathcal{N}_\phi})$ inductively, as it is key to our algorithm.

**Theorem 3.1** *If $\phi$ is an instantiation of attributes $E_1, \ldots, E_k$, the equivalence interval $I(F^\rho_{\mathcal{N}_\phi})$ can be computed if we know the equivalence interval $I(F^\rho_{\mathcal{N}_{\phi, e_{k+1}}})$ for every value $e_{k+1}$ of $E_{k+1}$:*

$$I(F^\rho_{\mathcal{N}_\phi}) = \bigcap_{e_{k+1}} \left\{ x : x + w_{e_{k+1}} \in I(F^\rho_{\mathcal{N}_{\phi, e_{k+1}}}) \right\}.$$

In our algorithm, we associate the node *node* with the equivalence interval $I[node] = I(F^\rho_{\mathcal{N}_\phi})$ if *node* is reached by path $\phi$. Theorem 3.1 states that we can compute this equivalence interval if we are given the equivalence interval of every child of *node*. Therefore, we can compute the equivalence interval of every node in the ODD $\mathcal{D}$ inductively, with the end conditions $I[1\text{-SINK}] = [\rho, \infty)$ and $I[0\text{-SINK}] = (-\infty, \rho)$.

To identify isomorphic sub–ODDs, we employ $n + 1$ caches in our algorithm, one for each $k = 0, \ldots, n$, where the $k$-th cache will store nodes at depth $k$. In each cache, nodes are indexed by their equivalence intervals. Given some path $\psi$ of length $k$, we check if there already exists some *node* in the $k$-th cache where $\log O(c \mid \psi) \in I[node]$. If this is true, the ODD $\mathcal{D}$ will be built such that the path $\psi$ also reaches *node*.

Algorithm 1 shows the procedure BUILD-ODD($\mathcal{N}, \rho$), which returns the root of the ODD $\mathcal{D}$ that represents the naive Bayes classifier $F^\rho_\mathcal{N}$, with respect to attribute order $(E_1, \ldots, E_n)$. After initializations, the ODD is built recursively by calling the procedure BUILD-SUB-ODD($k, v$), shown in Algorithm 2. This procedure returns the root of the sub–ODD $\mathcal{D}_\phi$ that represents the naive Bayes classifier $F^\rho_{\mathcal{N}_\phi}$, where $\phi$ is an instantiation of $E_1, \ldots, E_k$, and $v = \log O(c \mid \phi)$ is the prior log-odds of $\mathcal{N}_\phi$.

The next theorem gives us a theoretical upper bound on the number of nodes in the ODD $\mathcal{D}$, and the time complexity of Algorithm 1. It can be proved using Theorem 2.2; see Appendix A.

**Theorem 3.2** *The number of nodes in the ODD $\mathcal{D}$ built by Algorithm 1 is at most:*

$$\sum_{k=0}^{n} \min\{\|E_1, \ldots, E_k\|, \|E_{k+1}, \ldots, E_n\| + 1\}.$$

*If all attributes have at most $b$ values, the space complexity is $O(b^{n/2})$. Moreover, the time complexity of Algorithm 1 is $O(nb^{n/2})$.*

---

**Algorithm 1** BUILD-ODD($\mathcal{N}, \rho$): returns the root of the ODD $\mathcal{D}$ that represents the naive Bayes classifier $F^\rho_\mathcal{N}$, with respect to attribute order $(E_1, \ldots, E_n)$, where $\mathcal{N} = (C, \{E_1, \ldots, E_n\}, \log O(c), \{w_{e_i}\})$ is a naive Bayes network, and $\rho$ is the threshold.

---
1-SINK ← CREATE-NODE()
$I[1\text{-SINK}] \leftarrow [\rho, \infty)$
STORE-IN-CACHE($n$, 1-SINK)
0-SINK ← CREATE-NODE()
$I[0\text{-SINK}] \leftarrow (-\infty, \rho)$
STORE-IN-CACHE($n$, 0-SINK)
return BUILD-SUB-ODD($0, \log O(c)$)

---

**Algorithm 2** BUILD-SUB-ODD($k, v$): returns the root of the sub–ODD $\mathcal{D}_\phi$ that represents the naive Bayes classifier $F^\rho_{\mathcal{N}_\phi}$, where $\phi$ is an instantiation of $E_1, \ldots, E_k$, and $v = \log O(c \mid \phi)$ is the prior log-odds of $\mathcal{N}_\phi$. We define the following procedures as: CREATE-NODE() returns a newly-created node; FIND-IN-CACHE($j, x$) returns *node* in the $j$-th cache where $x \in I[node]$, or NIL if no such node exists; STORE-IN-CACHE($j$, *node*) stores *node* in the $j$-th cache, indexed by $I[node]$; ADD-CHILD(*node*, *child*, $l$) adds *child* as a child of *node*, with $l$ being the label of the edge; OFFSET($S, \delta$) returns $\{x : x - \delta \in S\}$.

---
node ← CREATE-NODE()
$I[node] \leftarrow (-\infty, \infty)$
**for all** values $e_{k+1}$ of $E_{k+1}$ **do**
  $v_{child} \leftarrow v + w_{e_{k+1}}$
  $child \leftarrow$ FIND-IN-CACHE($k + 1, v_{child}$)
  **if** $child =$ NIL **then**
    $child \leftarrow$ BUILD-SUB-ODD($k + 1, v_{child}$)
  ADD-CHILD($node, child, e_{k+1}$)
  $I[node] \leftarrow I[node] \cap$ OFFSET($I[child], -w_{e_{k+1}}$)
STORE-IN-CACHE($k$, *node*)
return *node*

---

Therefore, for a naive Bayes classifier with $n$ attributes, we are able to convert it into an ODD in time and space that is no more than exponential in $n/2$. This is significant both theoretically and practically compared to the brute–force method which is exponential in $n$. Hence, classifiers with up to 50 attributes can be handled in practice. However, as we will show in our experimental results, the actual time and space required by the algorithm is usually much less than the theoretical upper bound, showing promise for classifiers with even more attributes.

Finally, we also note that the actual number of nodes in the ODD will depend on the attribute order, and in the following section, we will suggest some ordering heuristics which perform well in practice.

| $n$ | $\|\mathbf{E}\|$ | Bound | Random | Desc. | Asc. |
|---|---|---|---|---|---|
| 10 | 1024 | 99 | 64 | 56 | 51 |
| 15 | 32768 | 518 | 347 | 270 | 263 |
| 20 | $1 \times 10^6$ | 3080 | 2032 | 1541 | 1531 |
| 25 | $3 \times 10^7$ | 16395 | 11968 | 8753 | 8740 |
| 30 | $1 \times 10^9$ | 98317 | 66160 | 50116 | 50100 |

Table 1: Experimental results of building ODDs that represent random naive Bayes classifiers.

## 4 Experimental Results

We now show experimental results on building ODDs for both random and real–world naive Bayes classifiers using Algorithm 1.

In the first part of our experiment, we build ODDs that represent random naive Bayes classifiers with binary attributes $\mathbf{E} = \{E_1, \ldots, E_n\}$, for different values of $n$. The prior log–odds and all the weights of evidence of the naive Bayes network take on random values, which are translated to the log–odds space from the uniform probability space. The threshold is set at $\rho = 0$, meaning $F_{\mathcal{N}}^{\rho}(\mathbf{e}) = 1$ iff $Pr(c \mid \mathbf{e}) \geq 0.5$. We generate 100 random classifiers for each $n$, and the results are displayed in Table 1. The second column shows the number of instances, i.e., $\|\mathbf{E}\| = 2^n$, while the third column shows the theoretical upper bound on the number of nodes in the ODDs given by Theorem 3.2. The fourth column shows the average number of nodes in the ODDs built using 100 random attribute orders. As we can see, the number of nodes is on average about two–thirds of the bound. We also sort the attributes by the absolute differences of the weights of evidence, i.e., $|w_{e_i} - w_{\overline{e_i}}|$, where a larger absolute difference means the attribute $E_i$ has more evidential impact on the probabilities $Pr(c \mid \mathbf{e})$. The sizes of the ODDs built using the attribute orders with descending and ascending orders of evidential impact are shown in the fifth and sixth columns respectively. In both cases, the number of nodes is on average about half of the bound, an improvement over using random attribute orders.

In the second part of our experiment, we build ODDs that represent real–world naive Bayes classifiers. The naive Bayes networks are constructed by learning data obtained from the UCI Machine Learning Repository (*www.ics.uci.edu/~mlearn/MLRepository.html*). The threshold $\rho$ is also set at 0. The results are displayed in Table 2 for several networks. The second column shows $n$, the number of attributes in the classifier, while the third column shows $\|\mathbf{E}\|$, the number of instances. Note that many of the attributes in the networks are non–binary. The fourth column shows the theoretical upper bound on the number of nodes in the ODDs given by Theorem 3.2.[7] For each classifier, we build ODDs using 100 random attribute orders, plus the attribute orders with descending and ascending orders of evidential impact,[8] and the final column shows the least number of nodes among the ODDs built.

| Network | $n$ | $\|\mathbf{E}\|$ | Bound | Best |
|---|---|---|---|---|
| tic-tac-toe | 9 | 19683 | 247 | 58 |
| votes | 16 | 65536 | 774 | 396 |
| spect | 22 | $4 \times 10^6$ | 6153 | 609 |
| breast-cancer-w | 9 | $1 \times 10^9$ | 21117 | 4405 |
| hepatitis | 19 | $2 \times 10^{10}$ | 46794 | 9644 |
| kr-vs-kp | 36 | $1 \times 10^{11}$ | 917488 | 59905 |
| mushroom | 22 | $1 \times 10^{14}$ | $1 \times 10^8$ | 43638 |

Table 2: Experimental results of building ODDs that represent real–world naive Bayes classifiers.

The results we produce are very satisfactory, since for many of these classifiers, there is an intractable number of instances, yet we are able to build ODDs with at most 60000 nodes in the best cases. The number of nodes actually created are also often much less than the theoretical upper bound, even with a random attribute order, because many of the CPTs in the classifiers are sparse, i.e., filled with 0's and 1's. An example is the mushroom network. The time to run our algorithm is also relatively short, as it takes less than five seconds to build an ODD with about 60000 nodes.

We also note that although the sizes of the ODDs vary with the attribute orders, experimentally we find that for each classifier, the size of the ODD in the worst case is at most about twice the size of the ODD in the best case. Therefore, even with a random attribute order, we are able to build ODDs of reasonable size. In the future, we would like to explore other ordering heuristics. Our current method of sorting the attributes by ascending order of evidential impact gives us the best results in many, but not all cases.

Finally, our algorithm can also be augmented, without affecting its complexity, to generate *reduced ODDs* [1], which eliminate nodes whose outgoing edges all point to the same child. However, we find that after including this reduction step, the sizes of the ODDs decrease by less than 1% in many of the cases, and less than 5% in most of the cases. Therefore, we do not include this in our algorithm for simplicity of exposition.

---

[7] Because the bound varies with the attribute order if the attributes do not have the same cardinality (number of values), the bound displayed here is computed for the ODD with the best result obtained.

[8] For a non–binary attribute $E_i$, we use a measure [2] that computes the difference between the maximum and minimum weights of evidence, i.e. $\max_{e_i} w_{e_i} - \min_{e_i} w_{e_i}$.

## 5 Applications

Now that we have an algorithm for converting a naive Bayes classifier into an ODD, our goal in this section is to discuss the variety of applications enabled by the construction of such an ODD.

We first point out that ODDs are a tractable representation in the sense that they permit a number of operations on the functions they represent in time polynomial in their size, even though such operations are intractable in general. In particular, given two ODDs $\mathcal{D}$ and $\mathcal{D}'$ with respect to the same variable order, with sizes $s$ and $s'$, we can perform the following operations:

- Testing whether the ODDs are equivalent can be done in $O(s + s')$ time.

- Counting the number of instances mapped to 1/0 by ODD $\mathcal{D}$ (positive/negative instances) can be done in $O(s)$ time.

- Testing whether all positive/negative instances of ODD $\mathcal{D}$ satisfy some conjunction or disjunction of features (attribute/value pairs) can be done in $O(s)$ time.

- Conjoining or disjoining the ODDs $\mathcal{D}$ and $\mathcal{D}'$ can be done in $O(ss')$ time.

All of the above operations on ODDs are supported by standard packages such as the CU Decision Diagram Package (*vlsi.colorado.edu/~fabio/CUDD/cuddIntro.html*).
These operations, plus many others, can be combined to answer queries. For example, if we want to know the number of positive instances in the intersection of two classifiers, we can first conjoin the two classifiers and then perform a count operation.

The equivalence operation is one of the most important operations because if two Bayesian network classifiers are shown to be equivalent, we can use either network to model the domain for the purpose of classifying instances. This is helpful if we want to test whether simplifications to a Bayesian network, such as rounding off the parameters, change the classification of any instance. We can also check if adding another attribute will improve the classification ability of the network. For example, for the network in Figure 1, we may want to know if adding a particular new test will be beneficial in detecting pregnancy, i.e., given any set of results from the current tests, whether applying this new test may potentially support the presence or absence of pregnancy.

Moreover, we can use the equivalence operation to see if the classification outputs given by networks produced from different learning algorithms are the same when run over the same data set, since the networks will differ in the parameters and possibly the structure. We can also determine if adding some data samples will change the behavior of the classifier produced by any learning algorithm.

Another application of converting a naive Bayes classifier $F_\mathcal{N}^\rho$ into an ODD $\mathcal{D}$ is that we can effectively find the intervals identified by Theorems 2.1 and 2.3 as a side effect of our algorithm. This is due to the computation of the equivalence interval of every node in the ODD $\mathcal{D}$ by Algorithm 2. For example, we note that the equivalence interval $I(F_\mathcal{N}^\rho)$ identified by Theorem 2.1, which contains the allowable prior log–odds which will keep the classifier unchanged from $F_\mathcal{N}^\rho$, is equal to $I[root]$ if $root$ is the root of the ODD $\mathcal{D}$.

We can also find the intervals identified by Theorem 2.3, which contain the allowable weights of evidence of attribute $E_i$ which will keep the classifier unchanged from $F_\mathcal{N}^\rho$. However, in order to find these intervals, it is required that $E_i$ must come first in the attribute order used to build the ODD $\mathcal{D}$. In this case, if the node $child_{e_i}$ is the $e_i$ child of the root of the ODD $\mathcal{D}$, the equivalence interval $I[child_{e_i}] = I(F_{\mathcal{N}_{e_i}}^\rho)$ contains the allowable weight of evidence $w_{e_i}$ which will keep the classifier unchanged from $F_\mathcal{N}^\rho$.

Therefore, instead of enumerating all instances explicitly, we can find the intervals identified by Theorems 2.1 and 2.3 by building the corresponding ODD using Algorithm 1. The asymptotic time and space complexity is exponential only in $n/2$, where $n$ is the number of attributes, but as seen in Section 4, the actual time and space required are often much less.

## 6 Extending the Proposed Framework

We now discuss some important extensions to our framework, some of which are relatively straightforward, while others are subjects of future work.

**Non–binary class variables** Throughout this paper, we have made the restriction that the class variable $C$ of the naive Bayes network is binary. In the case that $C$ is non–binary, we may be interested in mapping an instance $\mathbf{e}$ to the value of $C$ which is most likely given $\mathbf{e}$. To handle this generalization, we need two extensions to our framework. First, an ODD will need to have multiple sinks corresponding to the different values of $C$, which is relatively straightforward conceptually and does not change complexity as long as the cardinality of $C$ is bounded. Second, our algorithm for building the ODD must be changed so that instead of computing the equivalence interval $I$ for every node in the ODD, we compute the equivalence region, whose dimension is $|C| - 1$.

**Beyond naive Bayes classifiers** We also plan on expanding our work beyond naive Bayes classifiers. In particular, we are interested in classifiers induced from Tree Augmented Naive Bayes networks (TANs) [4] and Augmented Naive Bayes networks (ANBs), which are both derivatives of naive Bayes networks. In these networks, directed edges are added between attributes to model the domain more accurately. Because of the added edges, our algorithm has to be modified, because the weights of evidence may no longer be independent of the instantiated attributes. The attributes must now be divided into groups, such that two attributes in different groups are independent given the class variable $C$. Then, the ODD is built with respect to an order of groups, where every node in the ODD branches on instantiations of variables in a group. If there are $x$ variables in each group, and a total of $y$ groups, the theoretical upper bound on the number of nodes in the ODD is $O(b^{xy/2}) = O(b^{n/2})$, in the case where all attributes have at most $b$ values.[9] Therefore, the space complexity remains the same.

The ultimate goal of our future work is to generalize our algorithm to build logical representations corresponding to classifiers induced by any Bayesian network, and bound their sizes using measures of the network and the attribute order, such as the tree width.

## 7 Conclusion

In this paper, we introduced an algorithm for converting a naive Bayes classifier into an ODD, and proved a theoretical upper bound on the number of nodes in the ODD, which is asymptotically much less than the number of instances. Our experimental results showed that for real-world classifiers, the ODDs built tends to have even much fewer nodes than the bound. For applications, we showed how we can use our ODD to tractably reason about classifiers by applying a number of operations, such as testing for equivalence of two classifiers, in time linear in the size of the ODD. We also identified the range of allowable changes to a CPT in the network which keeps the current classifier unchanged. We believe this conversion from naive Bayes classifiers to a tractable logical representation are quite promising and helpful in practice, and plan on extending to general Bayesian network classifiers.

## Acknowledgments

This work has been partially supported by NSF grant IIS-9988543 and MURI grant N00014-00-1-0617.

---

[9]Of course, $x$ needs to be bounded because otherwise we will get an exponential branching factor.

## References

[1] Randal E. Bryant. Graph-based algorithms for Boolean function manipulation. *IEEE Transactions on Computers*, C-35(8):677–691, 1986.

[2] Hei Chan and Adnan Darwiche. A distance measure for bounding probabilistic belief change. In *Proceedings of the Eighteenth National Conference on Artificial Intelligence (AAAI)*, pages 539–545, Menlo Park, California, 2002. AAAI Press.

[3] R. O. Duda and P. E. Hart. *Pattern Classification and Scene Analysis*. John Wiley & Sons, New York, 1973.

[4] Nir Friedman, Dan Geiger, and Moises Goldszmidt. Bayesian network classifiers. *Machine Learning*, 29(2-3):131–163, 1997.

[5] Ashutosh Garg and Dan Roth. Understanding probabilistic classifiers. In *Proceedings of the Twelfth European Conference on Machine Learning (ECML)*, pages 179–191, Berlin, Germany, 2001. Springer-Verlag.

[6] Finn Verner Jensen. *Bayesian Networks and Decision Graphs*. Springer-Verlag, New York, 2001.

[7] P. Langley, W. Iba, and K. Thompson. An analysis of Bayesian classifiers. In *Proceedings of the 1992 National Conference on Artificial Intelligence (AAAI)*, pages 223–228, Menlo Park, California, 1992. AAAI Press.

[8] Judea Pearl. *Probabilistic Reasoning in Intelligent Systems: Networks of Plausible Inference*. Morgan Kaufmann Publishers, San Mateo, California, 1988.

## A Proofs

**Proof of Theorem 2.1** From Equation 2, we have $\log O'(c \mid e) = \log O(c \mid e) + \delta$, where $\delta = \log O'(c) - \log O(c)$. We investigate both cases of $\delta$ being negative and positive:

- If $\delta$ is negative, for every $e$ such that $F_{\mathcal{N}}^{\rho}(e) = 0$, we must have $F_{\mathcal{N}'}^{\rho}(e) = 0$. On the other hand, for every $e$ such that $F_{\mathcal{N}}^{\rho}(e) = 1$, we still have $F_{\mathcal{N}'}^{\rho}(e) = 1$ iff $\log O(c \mid e) \geq \rho - \delta$. Therefore, the classifiers $F_{\mathcal{N}}^{\rho}$ and $F_{\mathcal{N}'}^{\rho}$ are the same iff $\alpha = \min_{e:\, F_{\mathcal{N}}^{\rho}(e)=1} \log O(c \mid e) \geq \rho - \delta$.

- If $\delta$ is positive, for every $e$ such that $F_{\mathcal{N}}(e) = 1$, we must have $F_{\mathcal{N}'}^{\rho}(e) = 1$. On the other hand, for every $e$ such that $F_{\mathcal{N}}^{\rho}(e) = 0$, we still have $F_{\mathcal{N}'}^{\rho}(e) = 0$ iff $\log O(c \mid e) < \rho - \delta$. Therefore, the classifiers $F_{\mathcal{N}}^{\rho}$ and $F_{\mathcal{N}'}^{\rho}$ are the same iff $\beta = \max_{\mathbf{e}:\, F_{\mathcal{N}}^{\rho}(\mathbf{e})=0} \log O(c \mid \mathbf{e}) < \rho - \delta$.

Therefore, the classifiers $F_{\mathcal{N}}^{\rho}$ and $F_{\mathcal{N}'}^{\rho}$ are the same iff $\delta \in [\rho - \alpha, \rho - \beta]$, which is equivalent to $\log O'(c) \in I(F_{\mathcal{N}}^{\rho}) = [\log O(c) + \rho - \alpha, \log O(c) + \rho - \beta]$. □

**Proof of Theorem 2.2** When we change the prior log–odds of a naive Bayes network, we induce a different classifier only when $\log O(c \mid \mathbf{e})$ for some instance $\mathbf{e}$ passes $\rho$, thereby changing the classification of $\mathbf{e}$ from 0 to 1 or from 1 to 0. Therefore, the number of distinct classifiers (including the current classifier) that can be induced by changing the prior log–odds is at most $\|\mathbf{E}\| + 1$, and is exactly $\|\mathbf{E}\| + 1$ if there does not exist two different instances $\mathbf{e}$ and $\mathbf{e}^*$ such that $\log O(c \mid \mathbf{e}) = \log O(c \mid \mathbf{e}^*)$. □

**Proof of Theorem 2.3** The proof is similar to that of Theorem 2.1, with $\log O'(c \mid \mathbf{e}) = \log O(c \mid \mathbf{e}) + \delta_{e_i}$, where $\delta_{e_i} = w'_{e_i} - w_{e_i}$ if $e_i \in \mathbf{e}$. □

**Proof of Theorem 2.4** The number of distinct classifiers (including the current classifier) that can be induced by changing all weights of evidence of attribute $E_i$ appears to be $(\|\mathbf{E} - E_i\| + 1)^{\|E_i\|}$ at first glance, because from Theorem 2.2, we know that $\|\mathbf{E} - E_i\| + 1$ distinct classifiers can be induced by changing the prior log–odds of the new network $\mathcal{N}_{e_i}$, which is obtained by removing attribute $E_i$ from $\mathcal{N}$, and for every value $e_i$ of $E_i$, the classifier $F_{\mathcal{N}_{e_i}}^{\rho}$ can be equivalent to any of these distinct classifiers if its prior log–odds $\log O(c \mid e_i)$ can take on any value. However, this is true only if we can also change $\log O(c)$. This is not true if $\log O(c)$ cannot be changed, because of the restriction that among all weights of evidence $w_{e_i}$, at least one must be positive and at least one must be negative (unless all are zero), due to the fact that when going from one probability distribution to another, at least one probability must increase and at least one must decrease (unless all probabilities are the same). To find the actual maximum number of distinct classifiers, we have to solve the following analogous problem with $k = \|\mathbf{E} - E_i\|$ and $b = \|E_i\|$: given $S = \{0, 1, \ldots, k\}$, and $a \in S$, what is the number of permutations of $(a_1, \ldots, a_b) \in S^b$, if $(\bigvee_{i=1}^{b} a_i \geq a) \wedge (\bigvee_{i=1}^{b} a_i \leq a)$? The answer is $(k+1)^b - a^b - (k-a)^b$, and the maximum is $(k+1)^b - \lfloor k/2 \rfloor^b - \lceil k/2 \rceil^b$, attained when $a = \lfloor k/2 \rfloor$. □

**Proof of Theorem 3.1** Given instantiations $\phi$ and $\psi$ of attributes $E_1, \ldots, E_k$, the following statements are equivalent:

1. Classifiers $F_{\mathcal{N}_\phi}^{\rho}$ and $F_{\mathcal{N}_\psi}^{\rho}$ are the same.

2. $\log O(c \mid \psi) \in I(F_{\mathcal{N}_\phi}^{\rho})$.

3. For every $e_{k+1}$ value of $E_{k+1}$, the classifiers $F_{\mathcal{N}_{\phi, e_{k+1}}}^{\rho}$ and $F_{\mathcal{N}_{\psi, e_{k+1}}}^{\rho}$ are the same.

4. For every $e_{k+1}$ value of $E_{k+1}$, $\log O(c \mid \psi, e_{k+1}) \in I(F_{\mathcal{N}_{\phi, e_{k+1}}}^{\rho})$.

Moreover, due to the probability relation from Equation 1, we have:

$$\log O(c \mid \psi, e_{k+1}) = \log O(c \mid \psi) + w_{e_{k+1}}. \quad (7)$$

Therefore, the equivalence interval $I(F_{\mathcal{N}_\phi}^{\rho})$ can be computed if we know the equivalence interval $I(F_{\mathcal{N}_{\phi, e_{k+1}}}^{\rho})$ for every value $e_{k+1}$ of $E_{k+1}$, by finding the set of values that satisfy Equation 7 for every $e_{k+1}$, and we have $I(F_{\mathcal{N}_\phi}^{\rho}) = \bigcap_{e_{k+1}} \left\{ x : x + w_{e_{k+1}} \in I(F_{\mathcal{N}_{\phi, e_{k+1}}}^{\rho}) \right\}$. □

**Proof of Theorem 3.2** Because a node in the $k$-th cache is reached by some path $e_1, \ldots, e_k$, the number of nodes in the $k$-th cache can be no more than $\|E_1, \ldots, E_k\|$. We also know that a node in the $k$-th cache is the root of a sub–ODD that represents a naive Bayes classifier with attributes $E_{k+1}, \ldots, E_n$. Theorem 2.2 shows that at most $\|E_{k+1}, \ldots, E_n\| + 1$ distinct classifiers can be induced by changing the prior log-odds, and this number also bounds the number of nodes in the $k$-th cache, since we do not create duplicate nodes corresponding to isomorphic sub-ODDs. Therefore, the number of nodes in the $k$-th cache is at most $\min\{\|E_1, \ldots, E_k\|, \|E_{k+1}, \ldots, E_n\| + 1\}$. This proves that the number of nodes in the ODD is at most $\sum_{k=0}^{n} \min\{\|E_1, \ldots, E_k\|, \|E_{k+1}, \ldots, E_n\| + 1\}$, since there are $n + 1$ caches, with $k = 0, \ldots, n$. We can also easily see that if all attributes have at most $b$ values, the space complexity is $O(b^{n/2})$. Moreover, because the nodes in each cache are indexed by their equivalence intervals, we can find and store the nodes in each cache using binary search. Therefore, the time complexity of Algorithm 1 is $O(nb^{n/2})$. □